# Fingertip Detection: A Fast Method with Natural Hand

Jagdish Lal Raheja
Machine Vision Lab
Digital Systems Group, CEERI/CSIR
Pilani, INDIA
jagdish@ceeri.ernet.in

Karen Das
Dept. of Electronics & Comm. Engg.
Don Bosco University
Assam, INDIA
karenkdas@gmail.com

Ankit Chaudhary
Computer Vision Research Group
Dept. of Computer Science
BITS Pilani, INDIA
ankitc.bitspilani@gmail.com

**Abstract-** Many vision based applications have used fingertips to track or manipulate gestures in their applications. Gesture identification is a natural way to pass the signals to the machine, as the human express its feelings most of the time with hand expressions. Here a novel time efficient algorithm has been described for fingertip detection. This method is invariant to hand direction and in preprocessing it cuts only hand part from the full image, hence further computation would be much faster than processing full image. Binary silhouette of the input image is generated using HSV color space based skin filter and hand cropping done based on intensity histogram of the hand image.

## 1. Introduction

Hand Gesture Recognition (HGR) is a very popular and effective way used for the human machine communication. It has been used in many applications including embedded systems, vision based systems and medical applications. In HGR, fingertip detection is an important part if image base models are being used. HGR systems face many problems in skin segmentation due to luminance and intensity in images. The fingertips detection models mostly have assumption about the hand direction; this restricts the natural expression of humans. Processing time is another key factor in image based processing algorithms.

Here we are focusing on direction invariant fingertip detection of natural hand with real time performance. This work requires no glove, sensors or color strips to detect the fingertips. The only assumption is that user will show the hand to system, facing the palm to the system while the direction of hand is not restricted. User is free to show hand in any direction as naturally hands move. This paper also present a figure cropping method based on hand size, which will fasten the further process as the processing pixels would be reduced after cropping. This paper is the extended version of our work published as [13]. Figure 1 describes the block diagram of the process.

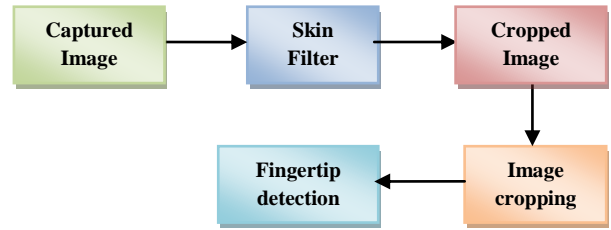

Figure 1. Algorithm Flow for the Fingertip Detection Method

## 2. Background work

A lot of work has been done in this area for dynamic hand gesture recognition using fingertip detection. A survey on fingertip based methods could be found out in [3]. There are several limitations in existing approaches. Garg [5] used 3D images in his method to recognize the hand gesture, but this process is complex and also not time efficient. The Processing time is one of the very critical factors in real time applications [11]. Aznaveh [1] presented a RGB vector based method for skin detection in images.

Yang [19] analyses the hand contour to select fingertip candidates, then finds peaks in their spatial distribution and checks local variance to locate fingertips. This method is not invariant to the orientation of the hand. There are other methods, which are using directionally Variant templates to detect fingertips [8][15]. Few other methods are dependent on specialized instruments and setup like the use of infrared camera [9], stereo camera [20], a fixed background [4][12] or use of markers on hand. This paper describes a novel method of motion patterns recognition generated by the hand without using any kind of sensor or marker.

The fingertips detection should be near to real time if it is going to process video. Generally image based models work on pixel by pixel and do hand Segmentation and work only on region of interest. However most hand segmentation methods can't do a clearly hand segmentation under some conditions like fast hand motion, cluttered background, poor light condition [6]. If the hand segmentation is not valid, then detection of fingertips can be questionable. Researchers [9][10][16] used infrared camera to get a reliable segmentation. Few researchers [4][6][7][12] [17][18] in their work limit the degree of the background clutter, finger motion speed or light conditions to get a reliable segmentation. Raheja [14] and few others also used 3D mapping using specialized device like KINECT for hand segmentation. Some of fingertip detection methods can't localize accurately multidirectional fingertips. Researchers [2][4][12][17] assumed that the hand is always pointing upward to get precise localization.

## 3   Approach to fingertips detection

Video is the sequence of the image frames at a fixed rate. First of all Images would be captured with a simple camera in 2D continuously and would be processed one by one. An HSV color space based skin filter would be applied on the images for hand segmentation. An intensity based histogram would be constructed for the wrist end detection and image would be cropped so that resultant image would have only hand pixels. The fingertips would be detected in the cropped hand image and would be marked differently.

### 3.1 Skin Filter

A HSV color space based skin filter would be used on the current image frame for hand segmentation. The skin filter would be used to create a binary image with black background. This binary image would be smoothened using the averaging filter. There can be false positives, to remove these errors the biggest BLOB (Binary Linked Object) is considered as the hand and rest are background as shown in figure 2(a). The biggest BLOB represent hand coordinates '1' and '0' to the background. The filtered out hand image is shown in figure 2(b), after removing all errors. The only limitation of this filter is that the BLOB for hand should be the biggest one.

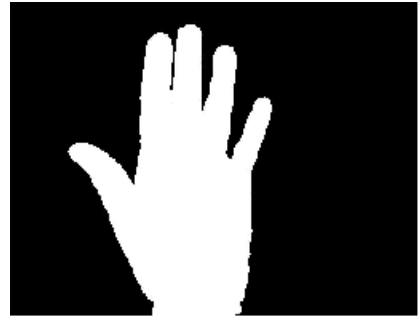

Figure 2(a). Biggest BLOB

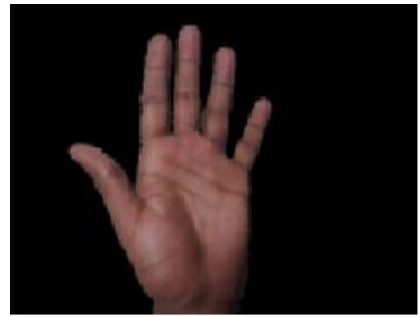

Figure 2(b). Hand after filtration

### 3.2 Wrist End Detection

To crop the image, we need to find out the direction of hand and wrist end. First of all an intensity histogram of the binary silhouette would be formed. Histograms functions are:

$$H_x = \sum_{y=1}^{n} imb(x, y)$$

$$H_y = \sum_{x=1}^{m} imb(x, y)$$

Here *imb* represents the binary silhouette and m, n represents the row and columns of the matrix *imb*.

To find the direction of hand, after a 4-way scan of image, we choose the maximum value of 'on' pixels coming out of all scans ('1' in the binary silhouette). it was noted that maximum value of 'on ' pixels represents wrist end and opposite end of this scan end would represent the finger end. Figure 3 shows the scanning process. The yellow bar showed in figure 3 corresponds to the first 'on' pixel in the binary silhouette scanned from the left direction scan. Similarly the green bar corresponds to right scan, red bar corresponds to down scan, and pink bar corresponds to up scan of 'on' pixels in the binary silhouette. Now, it is clear that red bar has higher magnitude than other bars. So we can infer that the wrist end is in downward direction of the frame and consequently the direction of fingers is in the upward direction. Here the direction from wrist to finger is known.

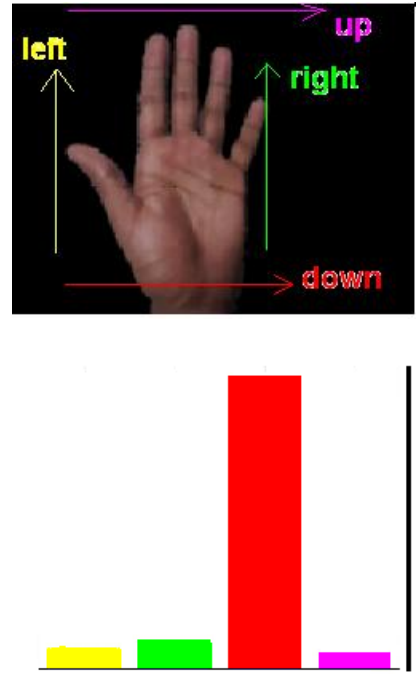

Figure 3. Image Scanning and Corresponding bars

### 3.3 Hand Cropping

Hand cropping minimizes the number of pixels to be taken into account for processing which leads to minimization of computation time. In the histogram generated in section 2.2, it was observed that at the point where wrist ends, a steeping inclination of the magnitude of the histogram starts as shown in figure 4, whose slope, m can be defined as:

$$m = \frac{y2 - y1}{x2 - x1}$$

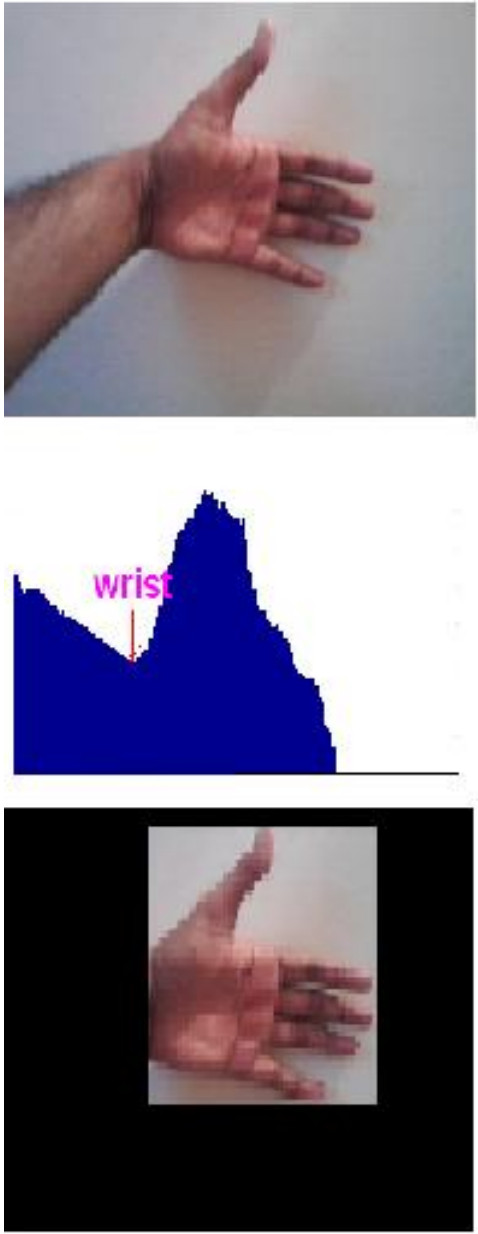

Figure 4. Hand Cropping Process: Images shown are (a) Initial Image, (b) Histogram of Binary Silhouette where wrist end is clearly detected, (c) Cropped Hand Image respectively.

The scan would start from the end where skin pixel was found and move till a inclination would be found. The points correspond to the first skin pixel scanning from other three sides were found, crop the image at that points. The equations for cropping the image are:

$$imcrop = \begin{cases} origin_{image}, & for\ Xmin < X < Xmax \\ & Ymin < Y < Ymax \\ 0, & elsewhere \end{cases}$$

Where imcrop represents the cropped image, Xmin, Ymin, Xmax, Ymax represent the boundary of the hand in the image.

Some results with processing steps for hand cropping are shown in figure 5. The arrows showed in the main frames indicate the direction of scan which was found from wrist end detection step. In all the histograms in figure 5 it is clear that at the wrist point, a steeping inclination starts in the scanning direction.

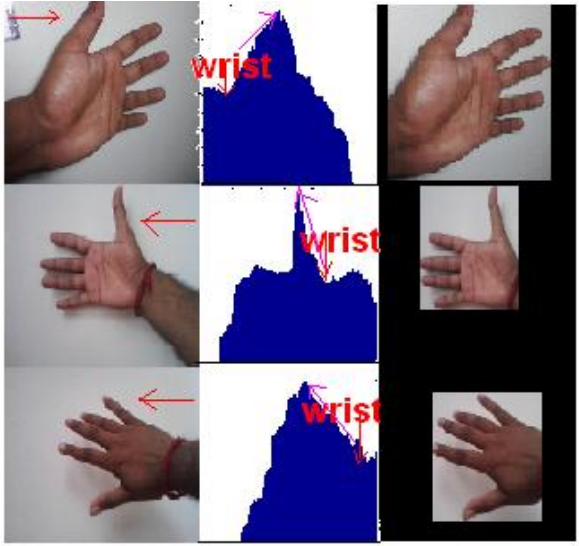

Figure 5. Results of Hand Cropping process from initial images

### 3.4 Fingertip Detection

At this point we have one smaller image which conation only skin pixels (hand gesture shape).  We will figure out fingertips in the cropped hand image. Start scan the cropped binary image from wrist to fingers end and calculate the number of pixels for each row or column based on the hand direction in up-down or left-right. Assign the intensity values for each pixel from 1 to 255 in increased manner from wrist to finger end in equal distribution.  So, each 'on' pixel on the edges of the fingers would be assigned a high intensity value of 255. Hence all the fingertips contain pixel value 255. The fingertip detection can be represented mathematically as,

$$pixel_{count}(y) = \sum_{X=Xmin}^{Xmax} imb(x,y)$$

$$modified_{image}(x,y) = round(x * 255 / pixel_{count}(y))$$

$$Finger_{edge}(x,y) = \begin{cases} 1 & if\ modified_{image}(x,y) = 255 \\ 0 & otherwise \end{cases}$$

Here $Finger_{edge}$ gives the boundary of the finger.

The line having high intensity pixel, is first indexed and check whether differentiated value lie inside a threshold, if it is then it represents a fingertip. The threshold value changes toward the direction of hand. That threshold can be set after the detection of the direction of hand to the finger which we already know. The results of fingertip detections are shown in figure 6 where detected pixels are marked with different color.

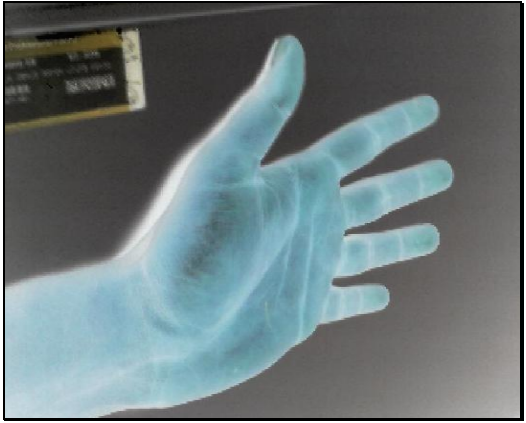

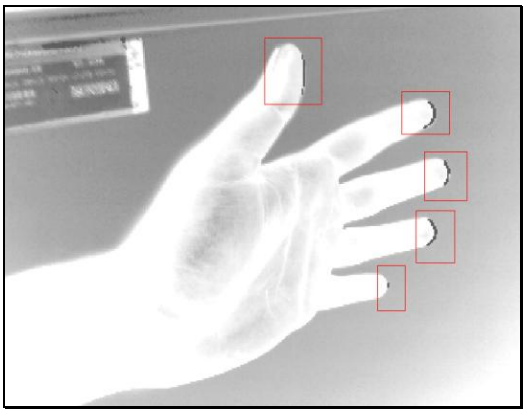

Figure 6. Results of fingertip detection in image frame

## 4. Conclusions

Here we have discussed a fast and efficient method for fingertips detection which will be used in our project 'Controlling the robot using hand gesture'. In this project user will pass controlling information to robot using hand gestures. The contribution by this paper is that user is free to show his hand in any direction, only palm should face the camera. The fingertips would be detected irrespective of user orientation. The Movement of user's finger will control the robot hand and its working, by moving hand in front of camera without wearing any gloves or markers.

## Acknowledgement

This research is being carried out at Central Electronics Engineering Research Institute (CEERI), Pilani, India as part of our project "Controlling the mechanical hand using hand gesture". Authors would like to thank Director, CEERI for his active encouragement and support.

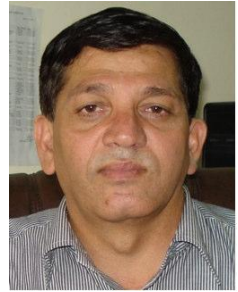

Dr. J.L.Raheja Received his M.Tech from Indian Institute of Technology, Kharagpur, INDIA and Ph.D. from Technical University of Munich, GERMANY. Currently he is senior scientist in Digital Systems Group at Central Electronics Engineering Research Institute (CEERI), Pilani, INDIA. His areas of research interest are Digital Image Processing, Embedded Systems and Human Computer Interface.

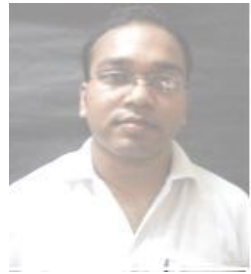

Karen Das received his M.Sc. in Electronics Science from Guwahati University, INDIA and M.Tech, in Electronics Engineering from Tezpur University, Assam. INDIA. His areas of research interest are Digital Image Processing, Artificial Intelligence, VLSI systems, and Digital Communication.

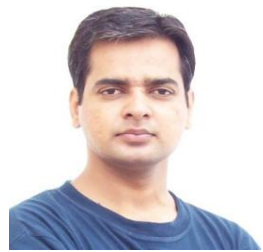

Ankit Chaudhary received his Master of Engineering degree in Computer Science & Engineering from Birla Institute of Technology & Science, Pilani, INDIA and currently working toward his Ph.D. in Computer Vision, from CEERI-BITS Pilani, INDIA. His areas of research interest are Computer Vision, Artificial Intelligence and Mathematical Computations.